\documentclass[letterpaper, 10 pt, conference]{ieeeconf} 
\IEEEoverridecommandlockouts      
\usepackage{times}
\usepackage[pdftex]{graphicx}
\usepackage{graphics}
\usepackage[cmex10]{amsmath}
\usepackage{url}
\usepackage{array}
\usepackage{tabularx}
\usepackage{multirow}
\usepackage{multicol}
\usepackage{booktabs}
\usepackage{epstopdf}
\usepackage{rotating}
\usepackage{csquotes}

\usepackage{amssymb}
\usepackage{amsmath}
\usepackage[all]{xy}    
\usepackage{ifthen}
\usepackage[usenames,dvipsnames]{color}
\usepackage{enumerate}
\usepackage{bm}
\usepackage{siunitx}
\usepackage{bbm}
\usepackage{xargs}                      
\usepackage[pdftex,dvipsnames]{xcolor}  
\usepackage[colorinlistoftodos,prependcaption,textsize=tiny]{todonotes}
\newcommandx{\unsure}[2][1=]{\todo[linecolor=red,backgroundcolor=red!25,bordercolor=red,#1]{#2}}
\newcommandx{\change}[2][1=]{\todo[linecolor=blue,backgroundcolor=white!25,bordercolor=blue,fancyline,#1]{#2}}
\newcommandx{\info}[2][1=]{\todo[linecolor=OliveGreen,backgroundcolor=OliveGreen!25,bordercolor=OliveGreen,#1]{#2}}
\newcommandx{\improvement}[2][1=]{\todo[linecolor=Plum,backgroundcolor=Plum!25,bordercolor=Plum,#1]{#2}}
\newcommandx{\thiswillnotshow}[2][1=]{\todo[disable,#1]{#2}}

\makeatletter
\let\NAT@parse\undefined
\makeatother
\usepackage[numbers]{natbib}

\newcommand{\secref}[1]{Section~\ref{#1}}
\newcommand{\tabref}[1]{Table~\ref{#1}}

\newcommand{\figref}[1]{Figure~\ref{#1}}

\newcommand{\probref}[1]{Problem~\ref{#1}}

\newcommand{\argmin}{\operatornamewithlimits{argmin}}

\newcommand{\myparagraph}[1]{\vspace{0.1in}\noindent\textbf{#1}}

\newboolean{draft-mode}
\setboolean{draft-mode}{true}
\newcommand{\sidenote}[1]{\ifthenelse{\boolean{draft-mode}}{\marginpar{\tiny\raggedright\textsf{\hspace{0pt}#1}}}{}}
\DeclareRobustCommand{\pynote}[1]{\ifthenelse{\boolean{draft-mode}}{\textcolor{green}{\textbf{PY: #1}}}{}}
\DeclareRobustCommand{\arnote}[1]{\ifthenelse{\boolean{draft-mode}}{\textcolor{blue}{\textbf{AR: #1}}}{}}
\DeclareRobustCommand{\nfnote}[1]{\ifthenelse{\boolean{draft-mode}}{\textcolor{red}{\textbf{NF: #1}}}{}}
\DeclareRobustCommand{\mbnote}[1]{\ifthenelse{\boolean{draft-mode}}{\textcolor{cyan}{\textbf{MB: #1}}}{}}

\newcommand{\bx}{{\bf x}}
\newcommand{\be}{{\bf e}}

\newcommand{\bw}{{\bf w}}
\newcommand{\bz}{{\bf z}}

\title{\LARGE \bf Realtime State Estimation with Tactile and Visual Sensing.\\Application to Planar Manipulation.
}
\author{\authorblockN{Kuan-Ting Yu$^1$, Alberto Rodriguez$^2$} \authorblockA{$^1$
    Computer Science and Artificial Intelligence Laboratory ---
    Massachusetts Institute of Technology\\
    $^2$ Mechanical
    Engineering Department --- Massachusetts Institute of Technology\\
    {\tt\small peterkty@csail.mit.edu}, {\tt\small
      albertor@mit.edu}} \thanks{This work was supported by
    NSF award [IIS-1427050] through the
    National Robotics Initiative and the Toyota Research Institute.}
    \\{\color{blue}Project website:} \url{mcube.mit.edu/push-est}\quad
    {\color{blue}Video:} \url{youtu.be/AL6weOfg25s}
}

\begin{document}

\maketitle


\begin{abstract}
Accurate and robust object state estimation enables successful object manipulation. Visual sensing is widely used to estimate object poses. However, in a cluttered scene or in a tight workspace, the robot's  end-effector often occludes the object from the visual sensor. The robot then loses visual feedback and must fall back on open-loop execution. 

In this paper, we integrate both tactile and visual input using a framework for solving the SLAM problem, incremental smoothing and mapping (iSAM), to provide a fast and flexible solution. Visual sensing provides global pose information but is noisy in general, whereas contact sensing is local, but its measurements are more accurate relative to the end-effector. 
By combining them, we aim to exploit their advantages and overcome their limitations. 
We explore the technique in the context of a pusher-slider system. We adapt iSAM's measurement cost and motion cost to the pushing scenario, and use an instrumented setup to evaluate the estimation quality with different object shapes, on different surface materials, and under different contact modes.
\end{abstract}


\section{Introduction}

\label{sec:introduction}


\begin{figure}
  \begin{center}
    \includegraphics[width=0.91\linewidth]{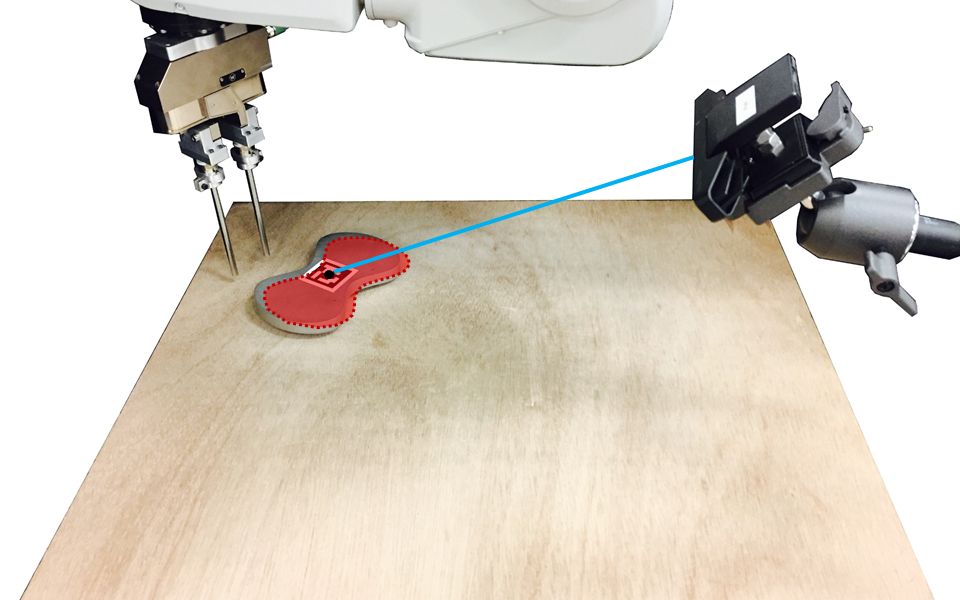}\vspace{0.1cm}
    \includegraphics[width=0.91\linewidth]{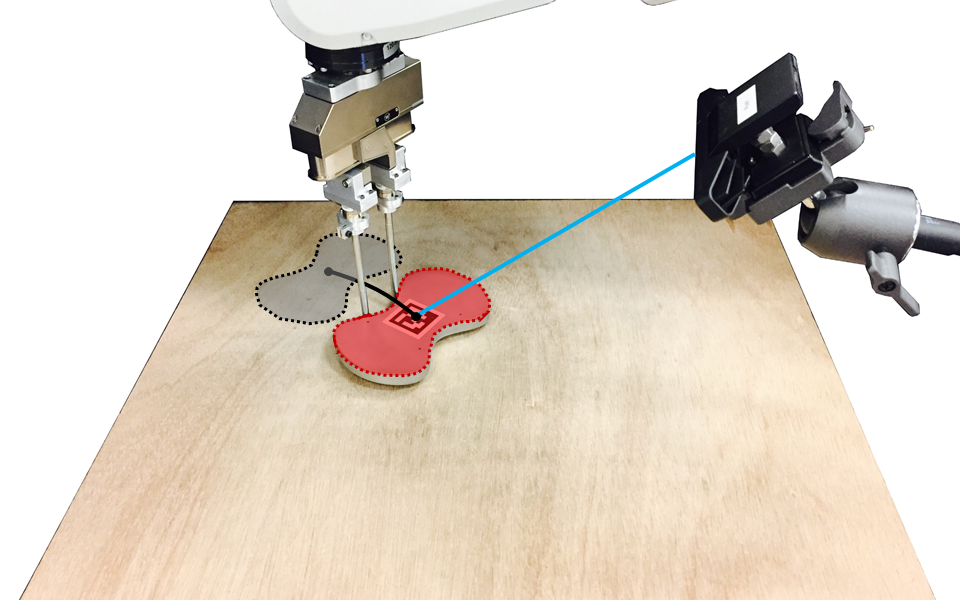}\vspace{0.1cm}
    \includegraphics[width=0.91\linewidth]{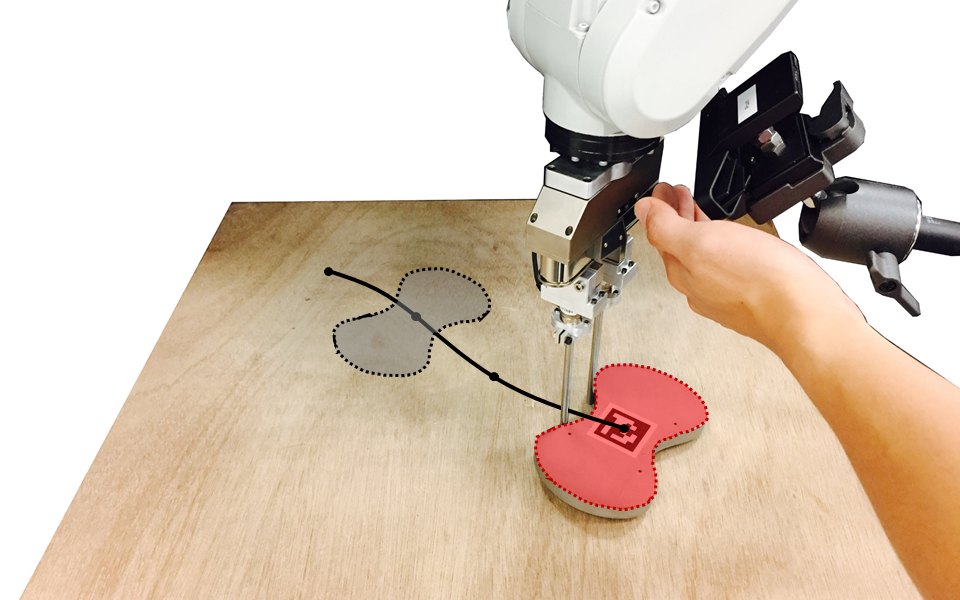}
  \end{center}
  \caption{Concept illustration. (top) Before the robot touches the object, the camera is able to track the object but with noticeable error due to imperfect calibration. (middle) After the robot makes contact with the object, the object pose is corrected based on the contact information. (bottom) During camera occlusion, the estimator can still keep track of the object while being pushed based on contact information. Red shape: current object pose estimate. Grey shape: object real pose in the last image. Black curve: object trajectory. Cyan: camera ray to object. }
  \label{fig:qualitative}
\end{figure}

We are interested in providing robot manipulators with the ability to track the state of a manipulation task in realtime with ordinary hardware.
Visual object tracking and detection have been widely studied \cite{zeng2016multi, zhang2012real, schmidt2015dart, jan_ICRA_2016} and can provide reliable estimates of object pose, especially in scenarios with limited occlusions.
However, it is characteristic of robotic manipulation that the robot, the gripper, or the surrounding clutter will ``get in the way" and occlude the object from the cameras.
%
In these scenarios, tactile sensing from distal sensors located at the end-effector can help track the state of an object. Therefore, new algorithms are needed to make sense of the high frequency but local information the sensors provide.

In this work, we describe a flexible state estimation framework that fuses in real time tactile and visual sensing. In previous work~\citep{yu2015shape}, we showed that it is possible to infer the shape and trajectory of a pushed object from a batch stream of tactile information. However, the performance was slow and not suitable for online tracking. In this paper, we explore the idea of combining \emph{visual sensing} for approximate global estimation with \emph{tactile sensing} for accurate interaction. We explore the algorithm in the context of a pusher-slider system~\citep{yu2016more}. The goal is to estimate 2D object poses in real time given intermittent contact measurements and unreliable visual measurements. The robot can push the object in various ways and with different contact modes (single-double contacts and sticking-sliding contacts). \figref{fig:qualitative} illustrates the idea.

We use incremental smoothing and mapping (iSAM)~\citep{Kaess08tro} as the underlying optimization framework because of its smoothing structure and the flexibility to fuse heterogeneous measurements and cost functions. 
The models we use to run the smoothing algorithm assume that we know the shape of the object and that the pressure distribution between the object and the ground is uniform. In practice, we do not have control over the pressure distribution in our experiments, but the algorithm still performs correctly.

Our system features are:
\begin{itemize}
    \item real-time estimation at 100~Hz;
    \item incorporation of contact physics in estimation;
    \item robustness to unreliable sensor input;
    \item acceptance of various ways of object pushing, which may involve changing the number of contacting fingers, or involve switching between sticking and sliding.
\end{itemize}

A key aspect of our system, enabled by the availability of tactile sensing, is that we do not require expensive complementarity programming of the sort that originates in classical contact problems (contact/no-contact or sticking/sliding) \citep{stewart1996implicit}. Relying on tactile and force measurements allows us to formulate the problem without complex hybrid dynamics, which speeds up the algorithm, while smoothing compensates for the possible sensor noise in detecting contact.



\section{Related Work}
\label{sec:relatedWork}
In this section, we discuss related literature from three aspects: A. state estimation for object manipulation, B. state estimation frameworks, and C. pushing mechanics.

\subsection{State estimation in object manipulation}
\citet{petrovskaya2011global} tackle the problem of global localization of a known and fixed object by touch. They apply a particle filter (PF) to fuse multiple point contact information. PFs can handle nonlinear systems and represent multiple modes. In practice, although pure touch-based localization is inspiring, in many real cases, it is often simple to add extra global sensors, e.g., cameras, which efficiently narrow down the search space. 

\citet{zhang2012application} also use PF to track an object during grasping acquisition with contact sensing patches on static fingers. They find that there is a particle depletion problem, which happens when the contact sensor yields very accurate measurements compared to those from cameras. The inability to fuse information of different accuracy scales is an inherent problem with particle filters.

\citet{koval2015} propose adding manifolds to resolve the problem of particle depletion. They use a probability distribution to keep track of binary variables representing the contact state. According to the variable, the filter uses a different set of dynamical constraints. 

Although contacting a surface is usually assumed to eliminate the uncertainty completely in the contact direction, it is only true if we use the exact contact point as the reference frame. If not, any physical extension from the reference point is not exactly precise and has different amounts of uncertainty.

\citet{li2015state} propose a contact graph that represents the transition of discrete contact states in order to let the contact state evolve according to physics.
In terms of computation, adding discrete variables will make the system unscalable due to a combinatorial number of contact modes of participating surfaces. In contrast, we use tactile sensing to determine contact states.


\citet{schmidt2015depth} focus on using depth pointclouds and contact constraints for state estimation. \citet{izatt2016gelsight} also use both a depth sensor and a high-resolution touch sensor. \citet{hebert2011fusion} fuse both visual and contact sensors. However, they do not consider motion models of the object during frictional contact interaction. In comparison, we add the motion constraints in order to prevent noisy measurements introducing unphysical estimates.

In previous work \cite{yu2015shape}, we showed that it is possible to recover not only the pose but also the shape of an unknown object during pushing exploration. We use a batch nonlinear least squares approach to incorporate both contact measurement and motion model constraints. The result was slow and not applicable to an online setting. In this paper, we adopt a similar formulation for the more practical problem of tracking the pose of a known object in an online fashion by adding visual input.

\subsection{State Estimation Framework}
The Extended Kalman Filter (EKF) is a popular framework for online and realtime estimation \cite{thrun2005probabilistic,izatt2016gelsight, hebert2011fusion}. It linearizes a system so as to apply a Kalman Filter, designed for linear systems. One common drawback with this approach is that the linearization point is chosen as the current estimate of variables. As it is often off the ground truth, this can result in an inaccurate linearization, followed by an inaccurate estimation.

\citet{Kaess08tro} propose incremental smoothing and mapping (iSAM) to solve the above issues. iSAM can be viewed as an online nonlinear least-square optimization tool, where cost functions and variables for the optimization can be added during each time step and can update the current estimate of the variables and linearization points. The update is fast because it uses a QR-factorized matrix to represent the linearized cost functions and only updates a very small fraction of the matrix. Also, when applicable, it can exploit the sparsity of the constraints.

Smoothing algorithms tend to provide more accurate estimates than filtering, e.g., the Extended Kalman Filter and particle filter (PF), because they maintain all the cost functions (soft constraints) of multiple timesteps instead of the last step only, and find the optimal solution based on all of them. Keeping a history of constraints can avoid noisy measurements to cause jumpy estimates. The algorithms also update the linearization point at a later stage to avoid the inaccurate linearization issue in the Kalman Filter.


\subsection{Pushing mechanics}
Pushing is a complex mechanical process because it involves two or more simultaneous frictional interactions between pushers and the object, and between the object and the surface. In terms of accuracy, the frictional interactions between real materials are uncertain; the uncertainty is demonstrated in \citet{yu2016more} by analyzing a large experimental pushing dataset. They found that friction properties are difficult to characterize precisely, and they change based on many factors including location on the surface and sliding direction.

\citet{lynch1992manipulation} propose a simple analytical model with a closed form expression. It is based on the Limit Surface (LS) \citep{Goyal1991} for force-motion mapping, and an ellipsoid approximation of LS \citep{lee1991fixture} for fast computation. We use Lynch's motion model as the pushing motion model due to its simplicity. There are more advanced pushing models which can be easily plugged in. For example, \citet{zhou2017} use a convex polynomial to represent LS more accurately. \citet{bauza2016probabilistic} use a Gaussian process to learn a stochastic model directly from data without physical modeling.

\section{Example Problem: Pusher-Slider Object Pose Estimation}

We are concerned with the problem of estimating the pose of a rigid 2D object pushed on a table in real time. 
The interaction between object and pusher is observed with
periodicity and we use the subscript $t\in[1...T]$ to indicate the
corresponding timestamp along the trajectory.

\myparagraph{Object pose.} We estimate the object pose denoted by $\bx_t = {(x,y,\theta)}$.

\myparagraph{Visual input.} A visual input includes a 2D pose $\bw_t$, and a binary variable denoting whether it is available at time $t$. We need the latter because the camera is sometimes occluded, or the frame has not arrived. Note that the occlusion could be caused by things other than the robot itself, such as a human co-worker.

\myparagraph{Tactile input.} A tactile input $\bz_{t}={\bz_{t,i}}$ includes force experienced $(f_x,f_y)_{t,i}$ and finger position $(p_x,p_y)_{t,i}$ in 2D on finger $i$. Finger positions are derived from robot joint states. We use $D_{t,i}$ to represent whether finger $i$ at time $t$ is in contact or not by setting a constant threshold $\tau$ on the force received. That is,
\[
    D_{t,i}=
\begin{cases}
    1,& \text{if } \left \|(f_x,f_y)_{t,i}\right \| \geq \tau\\
    0,& \text{otherwise}.
\end{cases}
\]

\begin{figure}
  \begin{center}
    \includegraphics[width=0.9\linewidth]{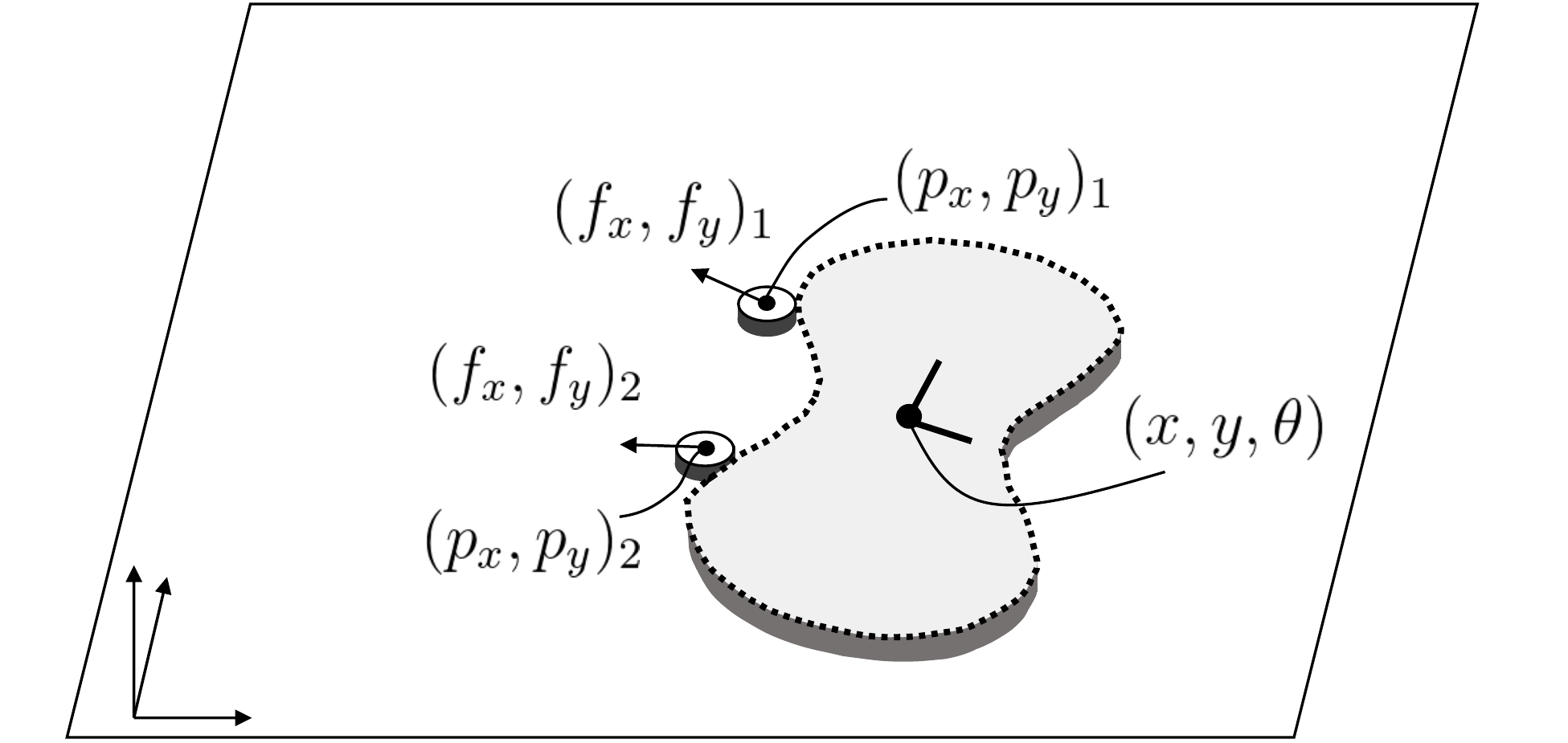}
  \end{center}
  \caption{Diagram for explaining the pushing state estimation problem.}
  \label{fig:diagram}
\end{figure}

We illustrate the above variables in \figref{fig:diagram}.

\section{Method}

\subsection{Applying iSAM}
Here we describe how we use iSAM in the context of the pushing state estimation problem. Refer to \cite{Kaess08tro} for the details of the iSAM algorithm. Below, we present the problem as solving a least squares problem; i.e., finding variables to minimize a cost function. 
Note that the variables and cost functions will be added and removed as time proceeds, in contrast to a batch optimization techniques. Note also that iSAM requires the assumption of a Gaussian noise model. We will test normality using real data in \secref{sec:experiment}.

\begin{figure}
  \begin{center}
    \includegraphics[width=0.9\linewidth]{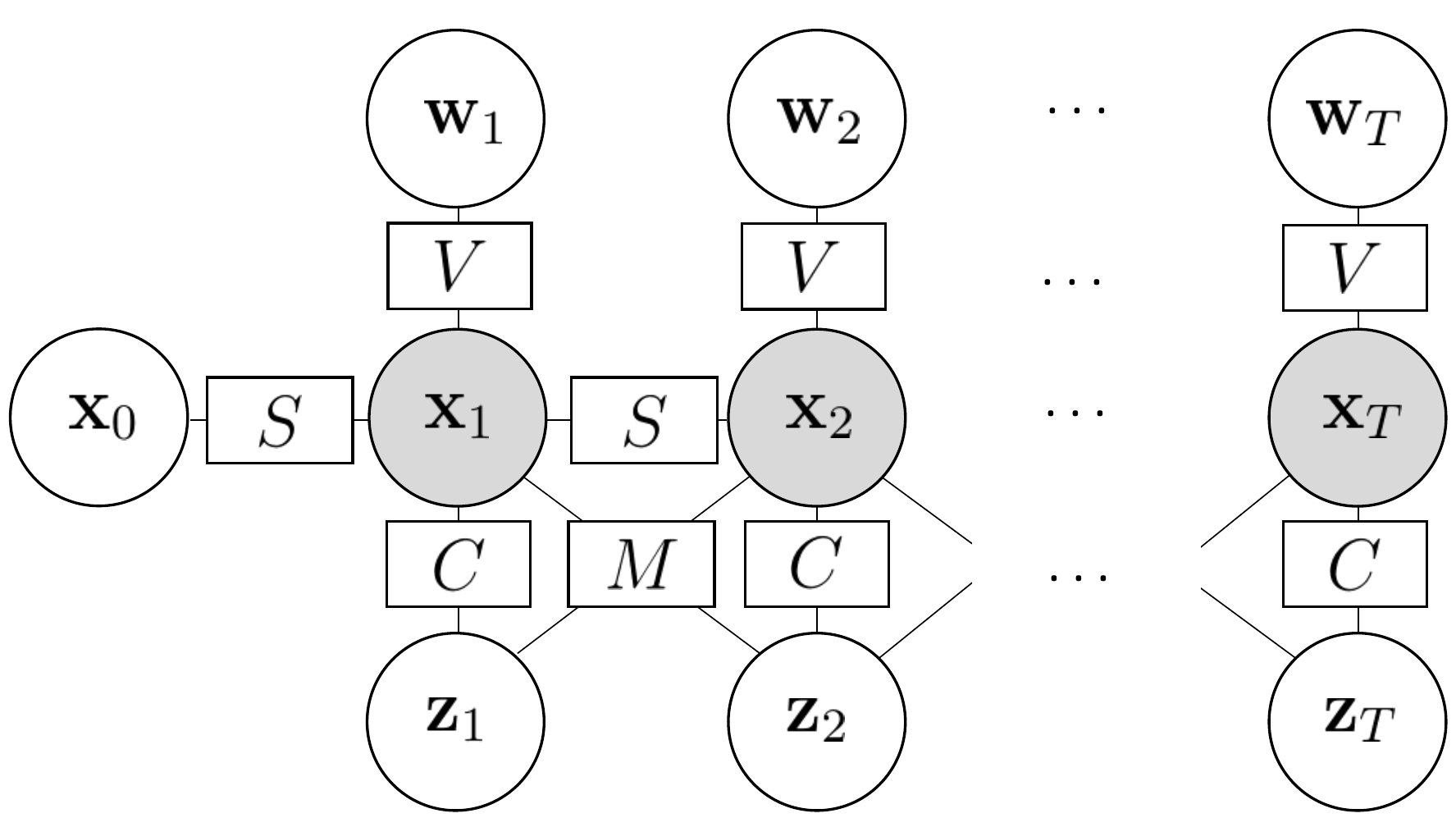}
  \end{center}
  \caption{A factor graph representation of the relationship between the variables and cost functions. The shaded circles represent the state variables to estimate, which are the object poses. The unshaded circles represent sensor measurements including camera inputs (\bw) and tactile inputs (\bz). The rectangles are the cost functions ($M, C, V, S$) that enforce soft constraints between the variables and measurements. }
  \label{fig:factor_graph}
\end{figure}

The overall cost function is a sum of four cost functions: 
\begin{itemize}
  \item the pushing motion cost $M$;
  \item the tactile measurement cost $C$; 
  \item the visual measurement cost $V$; 
  \item the stationary prior cost $S$. 
\end{itemize}
A factor graph in \figref{fig:factor_graph} shows the relationship between these cost functions. In summary, the overall least squares problem is 
\begin{equation}
\begin{aligned}
X^* = \argmin_X \sum_{t=1}^T 
&\left \| M(\bx_{t-1}, \bx_{t}, \bz_t, \bz_{t+1}) \right \|_\Lambda^2 \\
+&\left \| C(\bx_t, \bz_t) \right \|_\Gamma^2 \\
+&\left \| V(\bx_{t}, \bw_t) \right \|_\Upsilon^2  \\
+&\left \| S(\bx_{t}, \bx_{t-1}) \right \|_\Omega^2,
\end{aligned}
\end{equation}
where $X$ is a long vector formed by concatenating ${\bx_t}$'s, and $\left\|\be\right\|_\Sigma=\be^T\Sigma^{-1}\be$ computes squared Mahalanobis distance with covariance matrix $\Sigma$. The matrices $\Lambda$, $\Gamma$, $\Upsilon$, and $\Omega$ are the covariance matrices for the corresponding noise. We identify them from the measurement input and the ground truth. Some noises may not be constant over the work space, but in experiment they are pretty similar, so we average the noise levels across the testing space. If some measurement is missing due to physical limitations, we will remove the relevant cost functions; e.g., when the object is not in camera view, we remove the $V$ term. 

We always add a stationary prior because, in object manipulation, the object movement is almost zero in the time step of an estimation cycle, about 10 ms. From our experiments, this cost helps prevent the program from becoming underdetermined. 
On the other hand, it stabilizes the estimation result by filtering out jitters due to sensor noise.

Below we describe the four cost 
functions in detail.

\subsection{Physics-based pushing motion}

Lynch's pushing model \cite{lynch1992manipulation} is based on the following assumptions:
\begin{itemize}
    \item pushing at quasi-static speed;
    \item using ellipsoid limit surface approximation;
    \item assuming uniform friction between object and the surface and between object and pusher;
    \item assuming uniform pressure distribution between the object and the surface and between object and pusher.
\end{itemize}

Similar to previous work \citep{yu2015shape}, we can impose the motion model by using the following cost function:
\begin{equation}
M(\bx_{t-1}, \bx_{t}, \bz_t, \bz_{t+1}) = \left [\frac{v_x}{\omega} - c^2 \frac{F_x}{m},\frac{v_y}{\omega} - c^2 \frac{F_y}{m}\right]^T,
\end{equation}
where 
\begin{itemize}
 \item $v_x$ and $v_y$ are object velocities in $x$ and $y$ axes, derived from finite differences of $\bx_{t-1}$ and $\bx_{t}$;
 \item $\omega$ is angular velocity;
 \item $(F_x,F_y)$ is the total force, i.e., $\sum_i(f_x,f_y)_i$;
 \item $m$ is the total applied moment relative to current estimate of object center;
 \item $c$ is a scalar constant derived from the object pressure distribution.
\end{itemize}  All the variables are in the current object frame. 

Note that we do not need to distinguish between sticking or sliding because we sense the force acted on the object directly. By using limit surface representation, we can directly map the force acted on the object to the object velocity. Please refer to \cite{yu2015shape} for a more detailed derivation.

\subsection{Contact measurement}

\begin{figure}
  \begin{center}
    \includegraphics[width=0.5\linewidth]{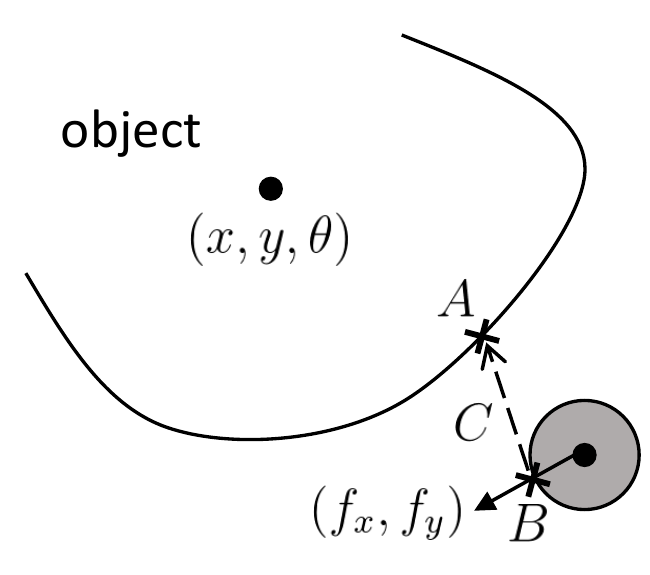}
  \end{center}
  \caption{Illustration of contact measurement cost. Point $A$ is the closest point from the object to the pusher. The object is at the pose described by the variables. Point $B$ is the contact point derived from the finger position and sensed force direction. Vector $C$ represents the distance between the two points that we want to minimize. }
  \label{fig:contact_measure}
\end{figure}

The contact measurement cost makes sure that when contact is detected, pusher and object are indeed in contact. 
The measurement cost is defined as the difference between the sensed contact point $B$ and the estimated closest point $A$ on the object with respect to the pusher contour: 
\begin{equation}
C(\bx_t, \bz_t) = A(\bx_t, \bz_t)-B(\bz_t).
\end{equation}
\figref{fig:contact_measure} illustrates the cost function and the two points. Note that this cost imposes not only that the contact point is right on the object boundary but also that the contact direction is correct.

\subsection{Visual measurement and stationary prior}

The visual measurement cost forces the pose estimate to be close to the visual input; the stationary prior forces the pose estimate to be close to the one from the last step. Both are implemented with a subtraction of  two inputs. That is,
\begin{equation}
\begin{aligned}
V({\bf x}_t,{\bf w}_t) & =  {\bf x}_t - {\bf w}_t, \text{ and} \\
S({\bf x}_{t}, {\bf x}_{t-1}) & =  {\bf x}_{t}-{\bf x}_{t-1}.
\end{aligned}
\end{equation}

Since the third element of the above subtractions is an angle, we need to wrap it into $[-\pi,\pi)$.


\section{Experiments}
\label{sec:experiment}

Our system estimates object poses at 100~Hz. It consumes visual inputs at 30~Hz and tactile inputs at 250~Hz.
We want to answer the following questions through our experiments:

\begin{itemize}
  \item Is contact measurement noise normally distributed? Since we assume a Gaussian noise model, we test the normality of the noise and find the covariance matrices for each cost function.
  \item Is iSAM a better parametric estimation framework than EKF in terms of estimation accuracy?
  \item How does each cost function contribute to form the final estimate? Which cost is able to correct which type of error?
  \item How does the estimation perform across different shapes? Is there any special geometry that is harder than others?
  \item How does the estimation perform across different surfaces?
  \item How fast can iSAM compute? Realtime computation is crucial to provide inputs for reactive control or planning.
\end{itemize}
More details about the experimental software and results are available at \url{mcube.mit.edu/push-est}.

\subsection{Hardware setup}

\begin{figure}
  \begin{center}
    \includegraphics[width=0.95\linewidth]{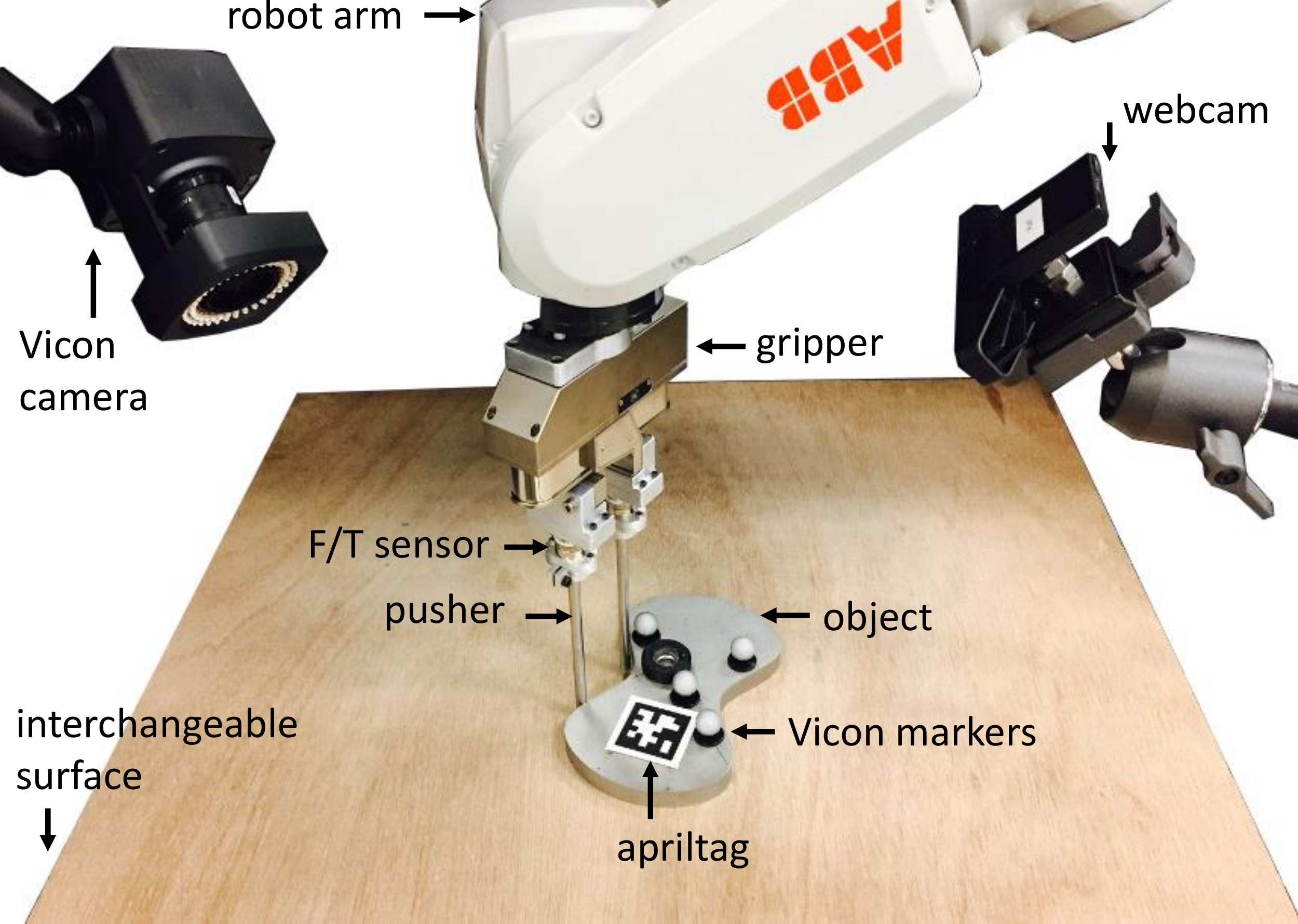}
  \end{center}
  \caption{Experimental hardware setup. }
  \label{fig:hardware_robot}
\end{figure}

To evaluate our method, we have an instrumented setup, as shown in \figref{fig:hardware_robot}: a 6 DOF industrial robotic manipulator equipped with two stiff cylindrical rods acting as a pusher. The setup is similar to that in our previous work where we collected an extensive pushing dataset \cite{yu2016more}.

\myparagraph{Robot.}
The system uses an ABB IRB 120 industrial robotic arm with 6 DOF to control precisely the position and velocity of its tool center point (TCP). The TCP moves at 60~mm/s in the experiments.

\myparagraph{Force sensing.}
We use two ATI Nano17 F/T sensors rigidly attached to the gripper to measure the reaction force from the object on the pusher. Since we only have force sensors but not contact sensors, we assume contact direction and sensed force direction is the same when we compute the contact measurement cost. In our experiments, we find them to be very close. We use the same constant $\tau$ to detect contact for all the experiments. 
Using only the force sensor allows the pusher to be very slim in appearance and strong mechanically.

\myparagraph{Pushers.}
The robot is equipped with two stiff cylindrical steel pushers, mounted on and perpendicular to the measurement plates of the force-torque sensor. The pusher has a length of 115~mm and diameter of 6.25~mm. 

\myparagraph{Objects.}
We use three objects, all water-jet cut in stainless steel. All objects are 13~mm thick. The friction coefficient between the pusher and the object is approximately 0.25, which was determined using a traditional variable slope experiment. A fiducial marker, Apriltag \citep{olson2011apriltag}, of 3~cm by 3~cm is stuck on the block to facilitate tracking by a webcam to obtain realistic visual object pose estimation input. The Apriltag system is able to detect occlusions. The objects are also instrumented with reflective markers and tracked with a Vicon motion tracking system for groundtruth. \tabref{tab:Dimensions} summarizes the objects that we experimented with. 

\myparagraph{Surface material.} 
We experiment with four surfaces: i) ABS, ii) Delrin, iii) plywood and iv) polyurethane (hardness 80A durometer). We have found different frictional characteristics in \citep{yu2016more}.

\begin{table}
  \caption{Objects used in the experiments. Physical properties.}
  \label{tab:Dimensions}
	\centering
		\begin{tabular}{|l|l|l|l|}
          \hline
          \bf Object & \texttt{rect1} & \texttt{ellip2} & \texttt{butter}\\
          \hline
          \bf Picture & \includegraphics[scale=0.3]{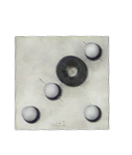} & \includegraphics[scale=0.3]{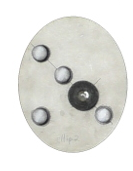} & \includegraphics[scale=0.3]{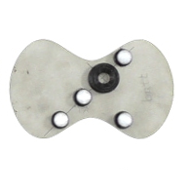} \\
          \hline
          \bf Mass (g) & 837 & 1110 & 1197\\
          \hline
          \bf Width (mm) & 90 & 105 & 95.3, 54.7\\
          \hline
          \bf Height (mm) & 90 & 130.9 & 156 \\
          \hline
        \end{tabular}
\end{table}

\begin{figure*}
  \begin{center}
    \includegraphics[width=1\linewidth]{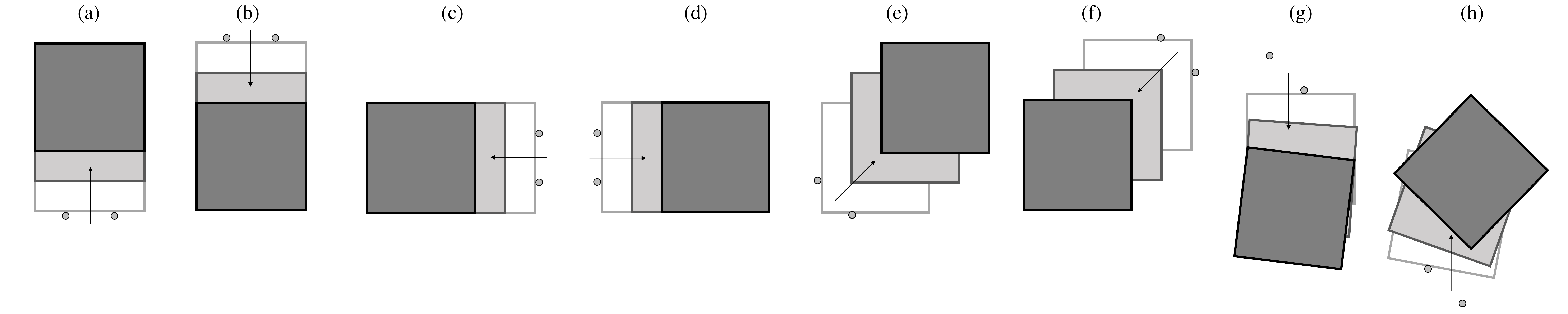}
  \end{center}
  \caption{Illustration of the standard testing pushing procedure. The webcam is located roughly at the right side of the block and looking left. (a)-(d) straight two finger pushes. (e)(f) corner two-finger pushes. (g)(h) one-finger pushes. Although some pushes might seem symmetric, the robot occludes the webcam in different ways.}
  \label{fig:push_seq_paper}
\end{figure*}

\myparagraph{Pushing procedure.} 
We use a pushing procedure to test the system. It covers several kinds of possible ways to push with two fingers: having contact and no contact; having sticking and sliding; having one finger and two fingers in contact; having occlusion or not. The procedure takes around 50~sec and is illustrated in \figref{fig:push_seq_paper}.

\myparagraph{Default configuration.} The default object for pushing is \texttt{rect1}, and the default surface is \texttt{plywood}. If below we do not specify the configuration, then we are using the default.

\myparagraph{Computation.} All computation was done on a laptop machine with Intel Core i7-3920XM CPU and 16~GB RAM.

\myparagraph{Baseline.} We use pure visual input without any filtering as our baseline method. When a visual input is not available at a time step, we use the latest available visual input.

\subsection{Noise characterization}

Since iSAM assumes a Gaussian noise model, we first want to test normality of the measurement noise models, and then find their covariance matrices. To find the error distributions, we evaluate the cost functions by using sensor measurements and the groundtruth object pose from Vicon. 

\begin{figure}
  \begin{center}
    \includegraphics[width=0.4\linewidth]{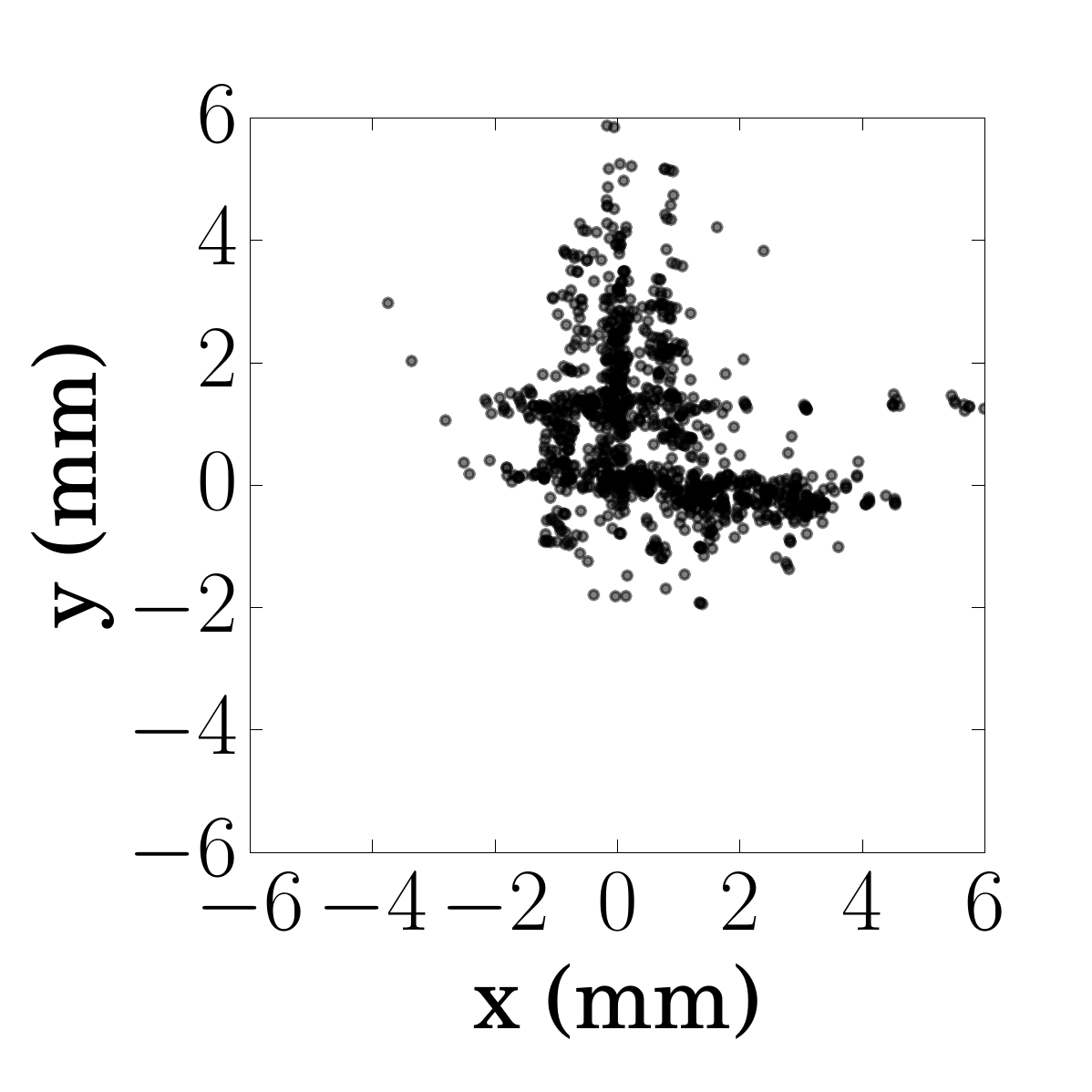}
    \includegraphics[width=0.4\linewidth]{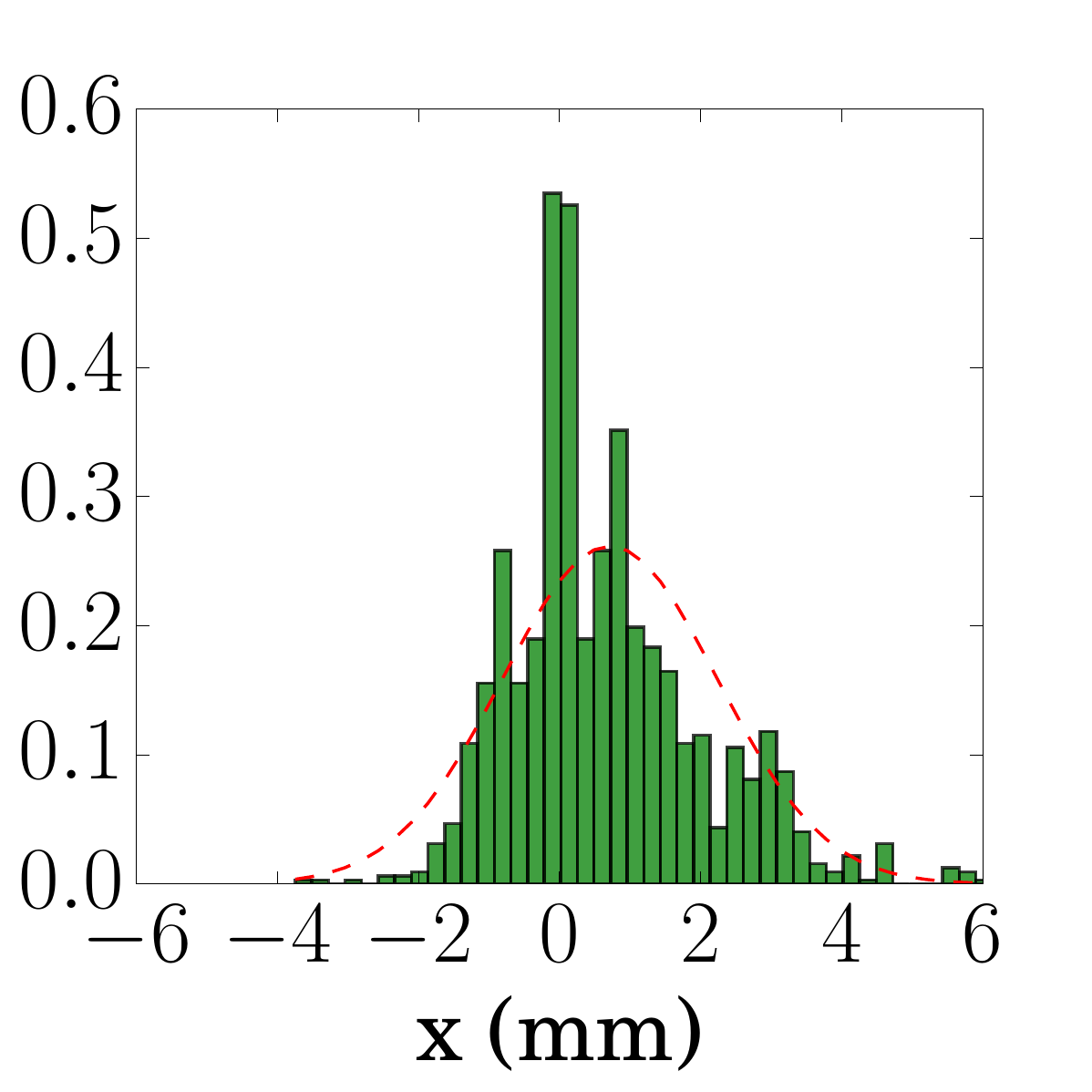} \\
    \includegraphics[width=0.50\linewidth]{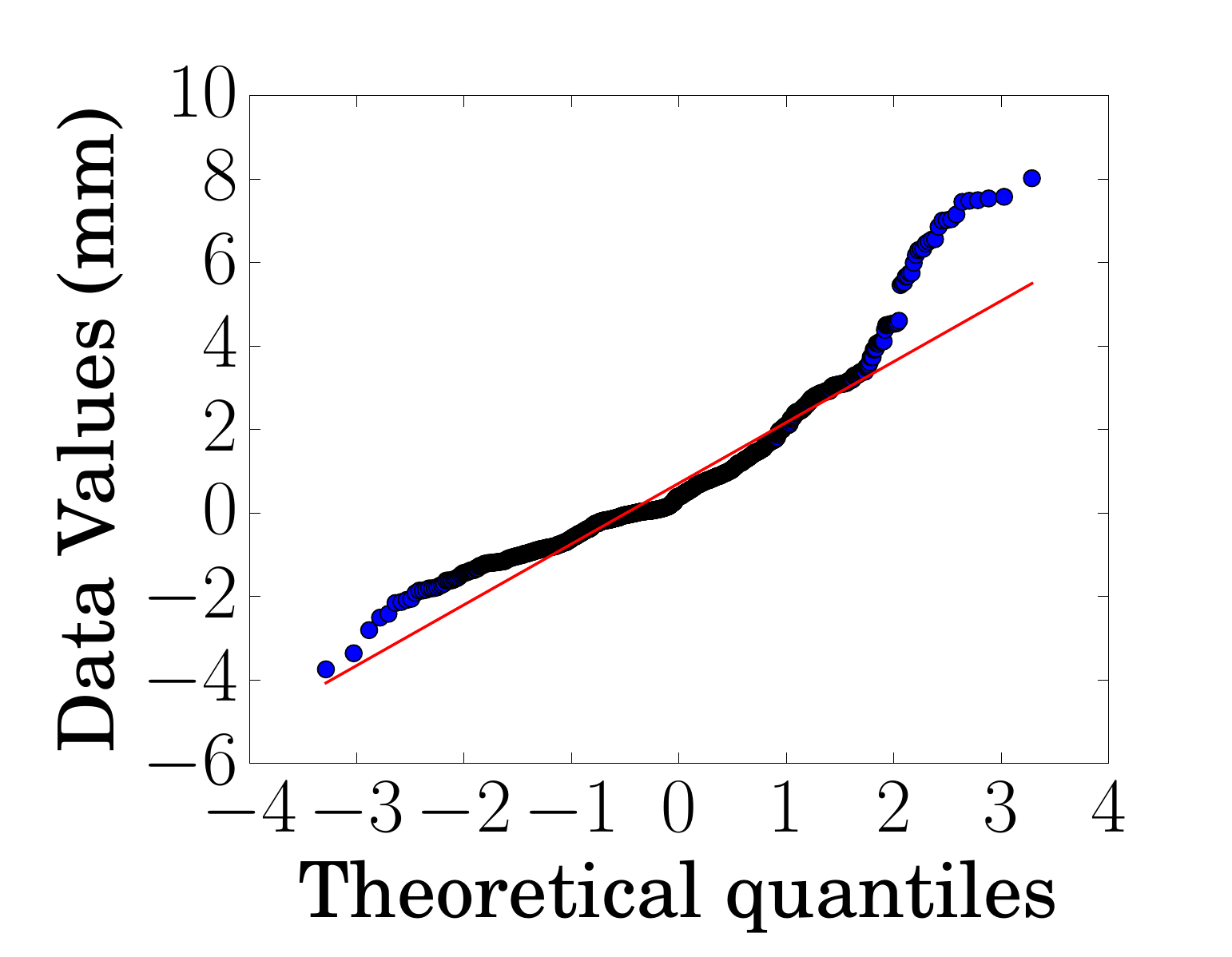}
  \end{center}
  \caption{Normality test of contact measurement error. Top left: scatter plot of the errors in the contact frame. Top right: histogram of the data projected to the first axis and a fitted Gaussian curve. Bottom: QQ plot in the first axis; the closer the data points to the straight line, the better the noise follows a Gaussian distribution.}
  \label{fig:noise_char}
\end{figure}

Our results confirmed that all the cost functions can be well approximated with a Gaussian distribution. We show the normality test for contact measurement in \figref{fig:noise_char}, which gives error distributions close to normal. While, in theory, an ideal contact should result in zero distance, in reality, any extension from the contact point will be imperfect. For example, the stiff pusher may deflect slightly when pushing an object such that the contact point given by the robot's kinematics does not match reality.

Given all the noises can be approximated as Gaussian distribution, we find the error covariance matrices using the groundtruth pose estimated from Vicon. 
In the following experiments, we use the same covariance matrices found with the default configuration because we find the parameters to be similar across different shapes and surfaces. In doing so, we can demonstrate the robustness of our estimation algorithm to variation.

\subsection{iSAM vs EKF, smoothing vs filtering}
In this section, we examine whether optimizing over a history of steps performs better than over one step like in EKF.  \tabref{tab:different_history} shows the result in terms of root mean squared error (RMSE) in translation and rotation.

We feed the same data with both visual and tactile information into EKF and iSAM using a different history length. We have two observations from the result:

\begin{itemize}
    \item EKF does improve the accuracy from pure visual input but is not as good as iSAM with multiple steps of history. Having a very short history makes the system more prone to abrupt sensor noise.
    \item Keeping longer history in iSAM, in general,  increases the accuracy, but the improvement becomes limited as the number of steps increase.
\end{itemize}

\begin{table}
  \caption{RMSE with different estimation methods. }
  \label{tab:different_history}
	\centering
	\begin{tabular}{|c|c|c|}
          \hline
          \bf Method & \bf Trans. & \bf Rot.   \\
          \bf  & \bf (mm) & \bf (deg) 
          \\ \hline
           Baseline  & 15.7$\pm$11.7 & 3.4$\pm$3.0 \\
           EKF   &  6.4$\pm$1.8 & 6.4$\pm$6.4 \\
           iSAM 1-step &  7.3$\pm$3.8 & 3.7$\pm$3.6 \\
           iSAM 100-step &  6.3$\pm$2.2 & 2.6$\pm$2.6 \\
           iSAM 200-step &  6.0$\pm$2.1 & 2.3$\pm$2.3 \\
           iSAM 1000-step &  6.1$\pm$2.1 & 1.8$\pm$1.8 \\
           \hline
    \end{tabular}
\end{table}

\subsection{Contribution of costs}

Visual input has its strengths and weaknesses.
It provides global information to guide the local tracking with motion and contact model. In our experiments, if the camera does not provide input for sufficiently long, we  lose track of the object. 
However, visual input alone is noisy and inaccurate. We see an average of a 15-mm translational error as shown in \tabref{tab:different_history}, which is due to calibration errors and occlusions. In that situation, contact feedback can help to refine the estimation. \figref{fig:different_shape} shows some examples where contact helps to refine the visual input.

\begin{figure*}
  \begin{center}
    \includegraphics[width=0.9\linewidth]{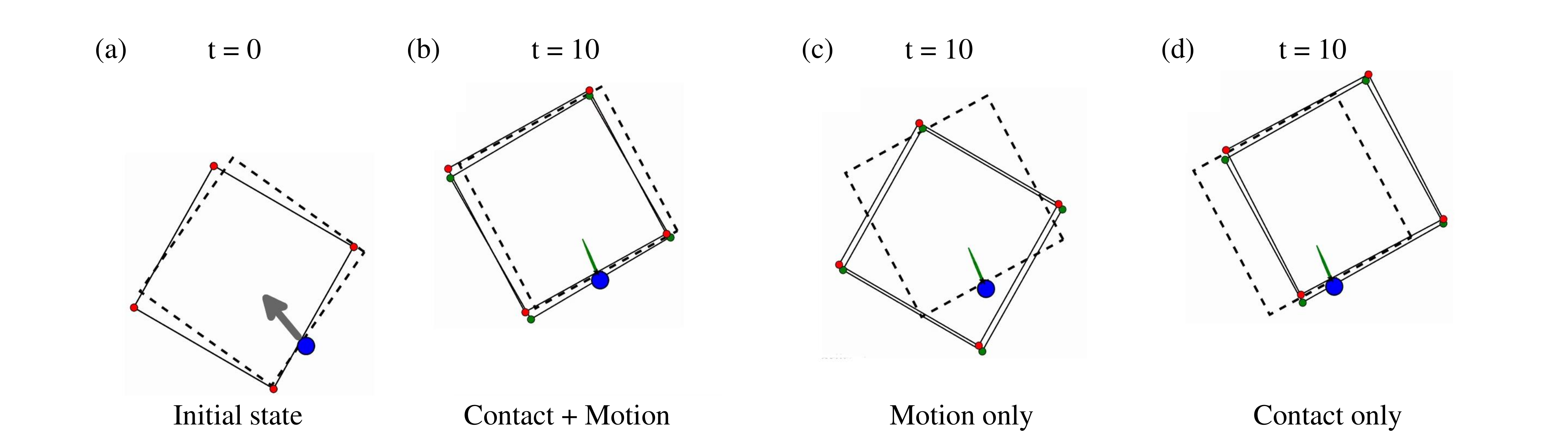}
  \end{center}
  \caption{Estimation results without Apriltag pose estimation due to occlusions and with different contact costs enabled. Red-dotted contour: current estimate of object pose. Green-dotted contour: estimate from the previous step. Dashed contour: groundtruth. (a) In the initial condition, the pusher pushes in the grey arrow direction. (b)(c)(d) are after 10 steps with the same push but with different contact costs enabled. (b) has both contact and motion costs. (c) has motion but not contact costs. (d) has contact but not motion costs. }
  \label{fig:exp_comparison}
\end{figure*}

Moreover, when there is no visual input for a sequence of time due to occlusion, we find that both the contact model and the motion model are important for accurate estimation. In \figref{fig:exp_comparison}, where visual inputs are not available, we show how contact measurement cost and motion cost contribute to good estimation. In (b), we see the estimation works well by using both costs. In (c), we use measurement cost but not contact cost, and the estimation fails because the motion prediction is very sensitive to the current estimate of the object pose. 
On the other hand, in (d), we use the contact model but not the motion model, and the estimated pose drifts perpendicularly to the contact normal.

\subsection{Varying object shapes}

\begin{figure}
  \begin{center}
    \includegraphics[width=0.95\linewidth]{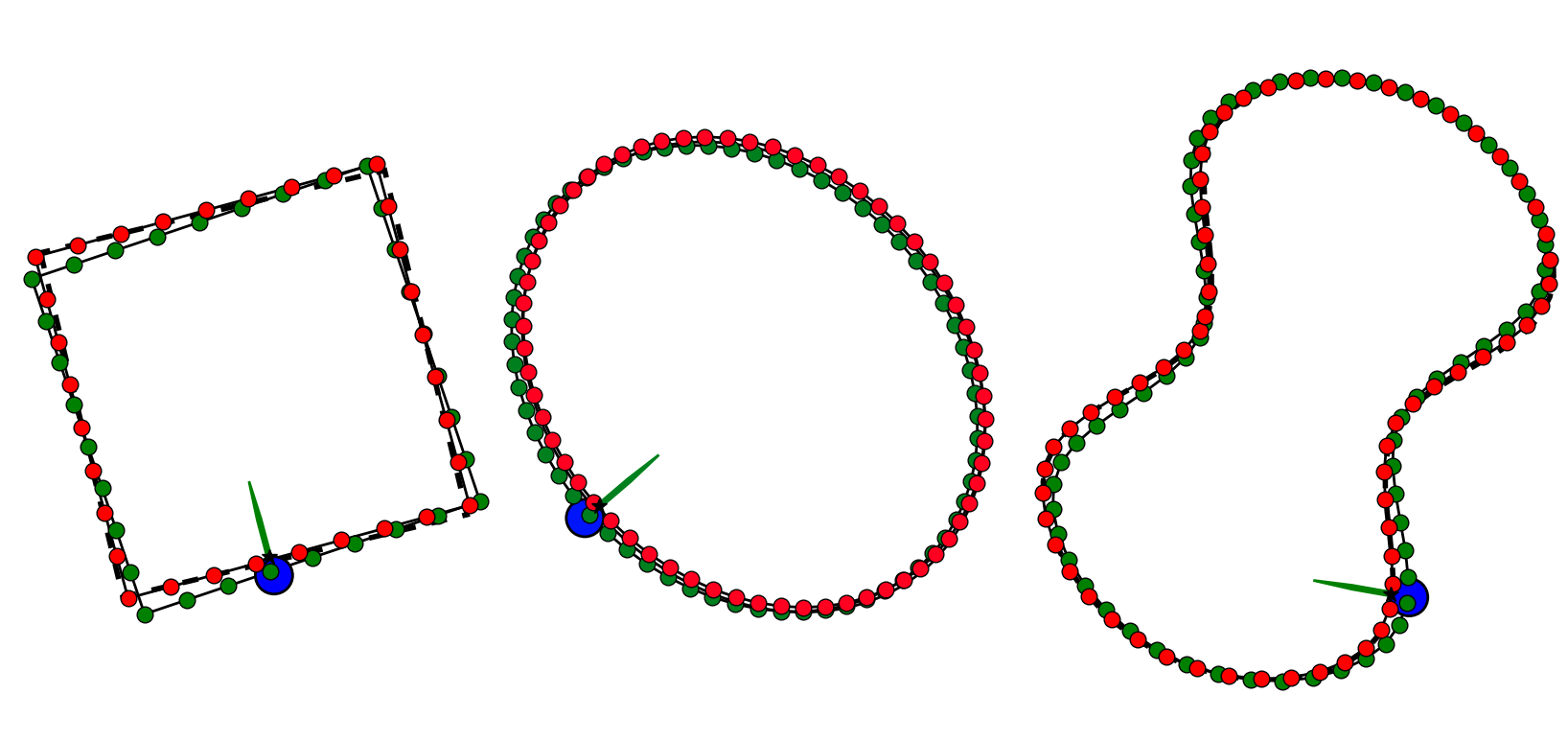}
  \end{center}
  \caption{Qualitative result of estimation. Here we emphasize the contribution of contact measurement cost. The green object contour: noisy pose input from visual pose detection. The red object contour: after iSAM update. The dashed object contour: groundtruth pose from Vicon. The blue circle: pusher position. The green arrow from pusher: sensed contact normal. Note the penetration between the pusher and the object from noisy input, which is corrected by the contact measurement.}
  \label{fig:different_shape}
\end{figure}

We first show that the cost functions can be applied to \texttt{rect1}, \texttt{ellip2}, and \texttt{butter} shapes. \figref{fig:different_shape} shows a qualitative result of validating contact measurement cost. The algorithm can be applied if the object can be approximated well as polygons and does not have small cavities where the pusher cannot enter.

We want to see if there are relationships between estimation accuracy and object shapes. \tabref{tab:different_shapes} shows the estimation accuracy with the three objects.
Although the standard pushing procedure may result in slightly different pushing interactions with the shape, we can still observe a general tendency. 
We observe that there are different error characteristic for different shapes. In general, for objects with smooth curves, i.e., \texttt{ellip2} and \texttt{butter}, the estimation error in rotation will be greater. We can reason it from a simple analysis: the measurement difference of nearby poses have a small gradient. In the extreme case, a circular object is ambiguous in all contact directions, so contact measurement will not be useful to distinguish object orientation.

Note that the baseline error of pushing \texttt{butter} is relatively high because the Apriltag was occluded more often by the robot and the Vicon markers.

\begin{table}
  \caption{RMSE with different shapes. }
  \label{tab:different_shapes}
	\centering
	\begin{tabular}{|c|r|r|r|r|}
          \hline
          
          \bf  & \multicolumn{2}{c|}{\bfseries Baseline}  & \multicolumn{2}{c|}{\bfseries iSAM}  \\ \hline
          
          \bf Shape & \bf Trans. & \bf Rot.  & \bf Trans. & \bf Rot. \\
          \bf  & \bf (mm) & \bf (deg) & \bf (mm) & \bf (deg)
          \\ \hline
           \texttt{rect1} & 15.7$\pm$11.7 & 3.4$\pm$3.0 & 6.0$\pm$2.1 & 2.3$\pm$2.3 \\
           \texttt{ellip2} & 16.7$\pm$14.6 & 4.0$\pm$3.3 & 7.3$\pm$4.9 & \textbf{5.7}$\pm$4.7 \\
           \texttt{butter} & 68.4$\pm$59.5 & 10.7$\pm$10.7 & 12.0$\pm$8.5 & \textbf{12.1}$\pm$10.4 \\ 
           \hline
    \end{tabular}
\end{table}

\subsection{Varying surfaces} 

We want to show that our solution works on different surfaces. In our previous work, each surface has different frictional properties \cite{yu2016more}. The variations include the variance of dynamic coefficient of friction and anisotropic/isotropic friction. In the experiment, we ensure the initial object pose is as similar as possible for fair comparison, so the pushing strokes are with respect to the initial pose of the object. \tabref{tab:different_surfaces} compares estimation accuracy on different surfaces and lists their dynamic coefficients of friction (DCoF). 

In most situations, the system greatly reduces the noise when fusing contact measurements. Only on the \texttt{delrin} surface is the rotation estimate worse than the baseline. We hypothesize that surface \texttt{delrin} has the smallest DCoF among all the surfaces, so force sensing is not as accurate compared to surfaces with higher DCoF. Therefore, after fusing it, the estimation in rotation become worse. We also see that \texttt{abs}, with low DCoF, has poorer accuracy in estimating rotation compared to the other surfaces.

Part of the accuracy improvement using contact comes from the fact that we use two fingers to push objects stably, reducing the uncertainties from pushing.


\begin{table}
  \caption{RMSE with different surfaces using rect1. }
  \label{tab:different_surfaces}
	\centering
	\begin{tabular}{|c|r|r|r|r|r|}
          \hline
          
          \bf  & & \multicolumn{2}{c|}{\bfseries Baseline}  & \multicolumn{2}{c|}{\bfseries iSAM}  \\ \hline
          
          \bf Surface & \bf DCoF & \bf Trans. & \bf Rot.  & \bf Trans. & \bf Rot. \\
          \bf  & & \bf (mm) & \bf (deg) & \bf (mm) & \bf (deg)
          \\ \hline
           \texttt{abs} & 0.16 & 14.3$\pm$10.9 & 7.8$\pm$7.5 & 5.6$\pm$2.7 & 3.7$\pm$3.6 \\
           \texttt{delrin} & 0.15 & 13.6$\pm$7.0 & 1.3$\pm$1.1 & 8.7$\pm$3.3 & \textbf{3.0}$\pm$2.9 \\
           \texttt{plywood} & 0.28 & 15.7$\pm$11.7 & 3.4$\pm$3.0 & 6.0$\pm$2.1 & 2.3$\pm$2.3 \\
           \texttt{pu} & 0.35 & 11.0$\pm$8.8 & 3.0$\pm$2.9 & 5.1$\pm$2.4 & 1.9$\pm$1.9 \\
           \hline
    \end{tabular}
\end{table}

\subsection{Timing}

The average time of computation is less than 1~ms for a 200-step history, including periodic relinearization. The linearization step is more time-consuming, averaging 28~ms ($\pm$ 21~ms) to linearize. The maximum linearization time was 70~ms, which corresponds to a pushing distance of 4.2~mm if we are pushing at 60~mm/s.

To ensure a constant processing rate, we choose to maintain a history of only 200 steps for all the results in this paper. We remove 100 nodes and related cost functions when the number of nodes reaches
300 steps. We do so because iSAM will relinearize every 100 steps, and removing nodes also requires relinearization. Therefore, we can save time by not relinearizing. The length can be chosen as a trade-off between computation speed and accuracy.

Note that the time reported above does not include visual input processing time. The Apriltag pose is tracked at 30~Hz. Using tactile sensor also helps with fast motion because contact sensors and robot poses are publishing at the higher rate of 250~Hz.
\section{Conclusion}
\label{sec:conclusion}

In this paper, we propose and demonstrate the use of iSAM for online estimation of object pose while pushing an object. 
Through extensive experiments, we understand how well the solution works in different task conditions, including different shapes, materials, and object interactions. 

We show that iSAM can give better accuracy than EKF by keeping a history of observations, while still being fast enough to allow realtime estimation.
By design, tactile sensing allows the system to distinguish between contact/no-contact and sticking/sliding without requiring expensive complementarity programming.


\myparagraph{Failure modes.}
In some challenging cases, the estimation loses track of the object. The main reasons are
\begin{itemize}
  \item the force sensor detects contact when there is no contact;
  \item the estimation does not receive good visual inputs for a long period of time.
\end{itemize}
A careful calibration of the tactile/force sensor will  help with the first problem. A conservative threshold for when there is contact will also help reduce damaging false contact positives. When contact is activated, a bad measurement will offset the estimate significantly because the contact measurement has high confidence.
The second problem can be alleviated if we plan for motions that facilitate visual perception.

\myparagraph{Limitations.}
This work has focused on tracking one object on a clean tabletop scenario. It may be possible to rely on occasional visual inputs to keep track of objects of interest in clutter and reason about the contact situation, but many details need to be addressed.
In general, an interaction of two or more objects without visual input is very challenging, even for humans. 

\myparagraph{Future Work.}
In the future, we would like to (1) test our estimation in the context of a pushing controller~\citep{hogan2016feedback} for reactive manipulation, which has been tested with accurate ground truth feedback from an external tracking system, and (2) generalize to 3D tasks such as prehensile manipulation~\citep{ChavanDafle2014,chavan2015prehensile,ChavanDafle2017}.  In both scenarios, state estimation is crucial to enable robot reactiveness and correct for dynamic, motion and sensor noise.

The proposed algorithm relies on the structure of a basic localization problem. There are many concepts and advances from the SLAM community, such as more complex data association schemes, that can be applied in  manipulation scenarios. A motivating example is what to do when a high-fidelity contact sensor like Gelsight~\citep{johnson2009retrographic} is available. Texture and salient geometric features can help with associating measurement with the object model. 


\bibliographystyle{IEEEtranN} 
{\footnotesize \bibliography{pushing-state-est}} 

\begin{thebibliography}{27}
\providecommand{\natexlab}[1]{#1}
\providecommand{\url}[1]{#1}
\csname url@samestyle\endcsname
\providecommand{\newblock}{\relax}
\providecommand{\bibinfo}[2]{#2}
\providecommand{\BIBentrySTDinterwordspacing}{\spaceskip=0pt\relax}
\providecommand{\BIBentryALTinterwordstretchfactor}{4}
\providecommand{\BIBentryALTinterwordspacing}{\spaceskip=\fontdimen2\font plus
\BIBentryALTinterwordstretchfactor\fontdimen3\font minus
  \fontdimen4\font\relax}
\providecommand{\BIBforeignlanguage}[2]{{%
\expandafter\ifx\csname l@#1\endcsname\relax
\typeout{** WARNING: IEEEtranN.bst: No hyphenation pattern has been}%
\typeout{** loaded for the language `#1'. Using the pattern for}%
\typeout{** the default language instead.}%
\else
\language=\csname l@#1\endcsname
\fi
#2}}
\providecommand{\BIBdecl}{\relax}
\BIBdecl

\bibitem[Zeng et~al.(2017)Zeng, Yu, Song, Suo, Walker~Jr, Rodriguez, and
  Xiao]{zeng2016multi}
A.~Zeng, K.-T. Yu, S.~Song, D.~Suo, E.~Walker~Jr, A.~Rodriguez, and J.~Xiao,
  ``Multi-view self-supervised deep learning for 6d pose estimation in the
  amazon picking challenge,'' \emph{ICRA}, 2017.

\bibitem[Zhang et~al.(2012)Zhang, Zhang, and Yang]{zhang2012real}
K.~Zhang, L.~Zhang, and M.-H. Yang, ``Real-time compressive tracking,'' in
  \emph{European Conference on Computer Vision}.\hskip 1em plus 0.5em minus
  0.4em\relax Springer, 2012.

\bibitem[Schmidt et~al.(2015{\natexlab{a}})Schmidt, Newcombe, and
  Fox]{schmidt2015dart}
T.~Schmidt, R.~Newcombe, and D.~Fox, ``Dart: dense articulated real-time
  tracking with consumer depth cameras,'' \emph{Autonomous Robots}, 2015.

\bibitem[Issac et~al.(2016)Issac, W{\"u}thrich, Garcia~Cifuentes, Bohg, Trimpe,
  and Schaal]{jan_ICRA_2016}
\BIBentryALTinterwordspacing
J.~Issac, M.~W{\"u}thrich, C.~Garcia~Cifuentes, J.~Bohg, S.~Trimpe, and
  S.~Schaal, ``Depth-based object tracking using a robust gaussian filter,'' in
  \emph{ICRA}.\hskip 1em plus 0.5em minus 0.4em\relax IEEE, May 2016. [Online].
  Available: \url{http://arxiv.org/abs/1602.06157}
\BIBentrySTDinterwordspacing

\bibitem[Yu et~al.(2015)Yu, Leonard, and Rodriguez]{yu2015shape}
K.-T. Yu, J.~Leonard, and A.~Rodriguez, ``{Shape and Pose Recovery from Planar
  Pushing},'' in \emph{IROS}, 2015.

\bibitem[Yu et~al.(2016)Yu, Bauza, Fazeli, and Rodriguez]{yu2016more}
K.-T. Yu, M.~Bauza, N.~Fazeli, and A.~Rodriguez, ``More than a million ways to
  be pushed. a high-fidelity experimental dataset of planar pushing,'' in
  \emph{IROS}, 2016.

\bibitem[Kaess et~al.(2008)Kaess, Ranganathan, and Dellaert]{Kaess08tro}
M.~Kaess, A.~Ranganathan, and F.~Dellaert, ``{iSAM}: Incremental smoothing and
  mapping,'' \emph{IEEE Trans. on Robotics (TRO)}, vol.~24, no.~6, pp.
  1365--1378, Dec. 2008.

\bibitem[Stewart and Trinkle(1996)]{stewart1996implicit}
D.~E. Stewart and J.~C. Trinkle, ``An implicit time-stepping scheme for rigid
  body dynamics with inelastic collisions and coulomb friction,''
  \emph{International Journal for Numerical Methods in Engineering}, vol.~39,
  no.~15, pp. 2673--2691, 1996.

\bibitem[Petrovskaya and Khatib(2011)]{petrovskaya2011global}
A.~Petrovskaya and O.~Khatib, ``Global localization of objects via touch,''
  \emph{IEEE Trans. on Robotics (TRO)}, vol.~27, no.~3, pp. 569--585, 2011.

\bibitem[Zhang and Trinkle(2012)]{zhang2012application}
L.~Zhang and J.~C. Trinkle, ``The application of particle filtering to grasping
  acquisition with visual occlusion and tactile sensing,'' in \emph{ICRA},
  2012.

\bibitem[Koval et~al.(2015)Koval, Pollard, and Srinivasa]{koval2015}
M.~Koval, N.~Pollard, and S.~Srinivasa, ``Pose estimation for planar contact
  manipulation with manifold particle filters,'' \emph{IJRR}, vol.~34, no.~7,
  June 2015.

\bibitem[Li et~al.(2015)Li, Lyu, and Trinkle]{li2015state}
S.~Li, S.~Lyu, and J.~Trinkle, ``State estimation for dynamic systems with
  intermittent contact,'' in \emph{ICRA}, 2015.

\bibitem[Schmidt et~al.(2015{\natexlab{b}})Schmidt, Hertkorn, Newcombe, Marton,
  Suppa, and Fox]{schmidt2015depth}
T.~Schmidt, K.~Hertkorn, R.~Newcombe, Z.~Marton, M.~Suppa, and D.~Fox,
  ``Depth-based tracking with physical constraints for robot manipulation,'' in
  \emph{ICRA}, 2015.

\bibitem[Izatt et~al.(2017)Izatt, Mirano, Adelson, and
  Tedrake]{izatt2016gelsight}
G.~Izatt, G.~Mirano, E.~Adelson, and R.~Tedrake, ``Tracking objects with point
  clouds from vision and touch,'' in \emph{ICRA}, 2017.

\bibitem[Hebert et~al.(2011)Hebert, Hudson, Ma, and Burdick]{hebert2011fusion}
P.~Hebert, N.~Hudson, J.~Ma, and J.~Burdick, ``Fusion of stereo vision,
  force-torque, and joint sensors for estimation of in-hand object location,''
  in \emph{ICRA}, 2011.

\bibitem[Thrun et~al.(2005)Thrun, Burgard, and Fox]{thrun2005probabilistic}
S.~Thrun, W.~Burgard, and D.~Fox, \emph{Probabilistic robotics}.\hskip 1em plus
  0.5em minus 0.4em\relax MIT press, 2005.

\bibitem[Lynch et~al.(1992)Lynch, Maekawa, and Tanie]{lynch1992manipulation}
K.~M. Lynch, H.~Maekawa, and K.~Tanie, ``Manipulation and active sensing by
  pushing using tactile feedback.'' in \emph{IROS}, 1992.

\bibitem[Goyal et~al.(1991)Goyal, Ruina, and Papadopoulos]{Goyal1991}
S.~Goyal, A.~Ruina, and J.~Papadopoulos, ``{Planar Sliding with Dry Friction
  Part 1. Limit Surface and Moment Function},'' \emph{Wear}, 1991.

\bibitem[Lee and Cutkosky(1991)]{lee1991fixture}
S.~H. Lee and M.~Cutkosky, ``Fixture planning with friction,'' \emph{Journal of
  Manufacturing Science and Engineering}, vol. 113, no.~3, 1991.

\bibitem[Zhou et~al.(2017)Zhou, Bagnell, and Mason]{zhou2017}
J.~Zhou, A.~Bagnell, and M.~Mason, ``A fast stochastic contact model for planar
  pushing and grasping: Theory and experimental validation,'' in \emph{RSS},
  2017.

\bibitem[Bauza and Rodriguez(2017)]{bauza2016probabilistic}
M.~Bauza and A.~Rodriguez, ``A probabilistic data-driven model for planar
  pushing,'' in \emph{ICRA}, 2017.

\bibitem[Olson(2011)]{olson2011apriltag}
E.~Olson, ``Apriltag: A robust and flexible visual fiducial system,'' in
  \emph{ICRA}, 2011.

\bibitem[Hogan and Rodriguez(2016)]{hogan2016feedback}
F.~Hogan and A.~Rodriguez, ``Feedback control of the pusher-slider system: A
  story of hybrid and underactuated contact dynamics,'' in \emph{WAFR}, 2016.

\bibitem[Chavan-Dafle et~al.(2014)Chavan-Dafle, Rodriguez, Paolini, Tang,
  Srinivasa, Erdmann, Mason, Lundberg, Staab, and Fuhlbrigge]{ChavanDafle2014}
N.~Chavan-Dafle, A.~Rodriguez, R.~Paolini, B.~Tang, S.~S. Srinivasa, M.~A.
  Erdmann, M.~T. Mason, I.~Lundberg, H.~Staab, and T.~A. Fuhlbrigge,
  ``{Extrinsic Dexterity: In-Hand Manipulation with External Forces},'' in
  \emph{ICRA}, 2014.

\bibitem[Chavan-Dafle and Rodriguez(2015)]{chavan2015prehensile}
N.~Chavan-Dafle and A.~Rodriguez, ``Prehensile pushing: In-hand manipulation
  with push-primitives,'' in \emph{IROS}, 2015.

\bibitem[{Chavan-Dafle} and Rodriguez(2017)]{ChavanDafle2017}
\BIBentryALTinterwordspacing
N.~{Chavan-Dafle} and A.~Rodriguez, ``{Sampling-based Planning of In-Hand
  Manipulation with External Pushes},'' in \emph{ISRR}, 2017. [Online].
  Available: \url{http://arxiv.org/abs/1707.00318}
\BIBentrySTDinterwordspacing

\bibitem[Johnson and Adelson(2009)]{johnson2009retrographic}
M.~K. Johnson and E.~H. Adelson, ``Retrographic sensing for the measurement of
  surface texture and shape,'' in \emph{CVPR}, 2009.

\end{thebibliography}

\end{document}